\documentclass[sigconf]{acmart}
\AtBeginDocument{%
  }

\copyrightyear{2026}
\acmYear{2026}

\setcopyright{cc}
\setcctype{by}

\acmConference[MM '26]
  {Proceedings of the 34th ACM International Conference on Multimedia}
  {November 10--14, 2026}
  {Rio de Janeiro, Brazil}

\acmBooktitle{Proceedings of the 34th ACM International Conference on
Multimedia (MM '26), November 10--14, 2026, Rio de Janeiro, Brazil}

\acmDOI{10.1145/3767308.3836003}

\acmISBN{979-8-4007-2213-4/2026/11}

\usepackage{multirow}
\usepackage{flushend}
\usepackage{balance}

\begin{document}

\title{Open-Set Visual Text Forensics via Sparse-Constraint Rectified Flow}

\author{Jiangling Zhang}
% \affiliation{%
%   \department{VCIP \& TMCC \& DISSec\\
%     College of Computer Science}
%   \institution{Nankai University}
%   \city{Tianjin}
\affiliation{%
  \department[0]{VCIP \& TMCC \& DISSec}
  \department[1]{College of Computer Science}
  \institution{Nankai University}
  \city{Tianjin}
  \country{China}}
\email{340056@whut.edu.cn}

\author{Shuxuan Gao}
\affiliation{%
  \department{School of Computer Science and Artificial Intelligence}
  \institution{Wuhan University of Technology}
  \city{Wuhan}
  \country{China}}
\email{shuxuan.gao-353758@whut.edu.cn}

\author{Zeyu Chen}
% \affiliation{%
%   \department{VCIP \& TMCC \& DISSec\\
%     College of Computer Science}
%   \institution{Nankai University}
%   \city{Tianjin}
\affiliation{%
  \department[0]{VCIP \& TMCC \& DISSec}
  \department[1]{College of Computer Science}
  \institution{Nankai University}
  \city{Tianjin}
  \country{China}}
\email{chenzeyu@mail.nankai.edu.cn}

\author{Yichao Liu}
% \affiliation{%
%   \department{VCIP \& TMCC \& DISSec\\
%     College of Computer Science}
%   \institution{Nankai University}
%   \city{Tianjin}
\affiliation{%
  \department[0]{VCIP \& TMCC \& DISSec}
  \department[1]{College of Computer Science}
  \institution{Nankai University}
  \city{Tianjin}
  \country{China}}
\email{liuyichao@mail.nankai.edu.cn}

\author{Yu Zhou}
\authornote{Corresponding author.}
% \affiliation{%
%   \department{VCIP \& TMCC \& DISSec\\
%     College of Computer Science \&
%     College of Cryptology and Cyber Science}
%   \institution{Nankai University}
%   \city{Tianjin}
\affiliation{%
  \department[0]{VCIP \& TMCC \& DISSec}
  \department[1]{College of Computer Science \&
    College of Cryptology and Cyber Science}
  \institution{Nankai University}
  \city{Tianjin}
  \country{China}}
\email{yzhou@nankai.edu.cn}

% \renewcommand{\shortauthors}{Jiangling Zhang et al.}

% \renewcommand{\shortauthors}{Jiangling Zhang et al.}


%%
%% Short author list used only in page headers.
\renewcommand{\shortauthors}{Jiangling Zhang et al.}

\begin{abstract}

Rapidly evolving Generative AI enables sophisticated visual text manipulations that increasingly evade current forensic detectors. Existing discriminative models often overfit specific forgery patterns, limiting their generalization to unseen, open-set attacks. To address this challenge, we propose a generative detector that localizes tampering by estimating the local restoration cost required to align a query image with authentic visual-text statistics, rather than by learning forgery-specific decision boundaries. Specifically, we introduce \textbf{S}parse-\textbf{C}onstraint \textbf{R}ectified \textbf{F}low (SC-RF), a detector-oriented adaptation of Flow Matching for spatially sparse anomaly localization. We further mitigate data scarcity via self-supervised \textbf{Artifact Injection} and preserve high-frequency forensic traces using a pixel-space \textbf{F}orensic-\textbf{DiT}. Extensive experiments on three benchmarks show that our method achieves state-of-the-art performance, surpassing the runner-up by 3.2 and 4.8 percentage points in F1 and IoU, respectively. In particular, the proposed detector demonstrates strong zero-shot performance on challenging unseen text editing patterns. We further provide an auxiliary stress-test analysis showing that local harmonization produced by our model can weaken the statistical cues relied upon by existing detectors, offering a complementary vulnerability-analysis perspective.

\end{abstract}


% %%
% %% Keywords. The author(s) should pick words that accurately describe
% %% the work being presented. Separate the keywords with commas.
% \keywords{Do, Not, Use, This, Code, Put, the, Correct, Terms, for,


\maketitle

\section{Introduction}

Visual text in documents and natural scenes is an important carrier of semantic information. Recent Generative AI, particularly Diffusion Models \cite{rombach2022ldm, nichol2021improved, dhariwal2021diffusion}, can manipulate glyphs and strokes with few perceptual artifacts, creating substantial challenges for information security and forensic detection. Current forensic methods largely adopt a \textit{discriminative} paradigm, learning decision boundaries to separate authentic and tampered regions \cite{QuLLCPGJ23,WangXXWZZ22}. However, such models tend to overfit to specific forgery artifacts (e.g., compression traces or GAN fingerprints) rather than capturing intrinsic properties of authentic images \cite{wang2020cnn,Zhang2025LAMMViT}, leading to poor generalization under open-set forgeries. As generative techniques evolve rapidly, this closed-set paradigm becomes increasingly unsustainable. Instead of modeling diverse forgery patterns, we ask whether a detector can directly measure local deviations from authentic visual-text statistics. This offers a more robust formulation for open-set detection than learning forgery-specific decision boundaries.

\begin{figure}[t]
\centering
\includegraphics[width=\linewidth]{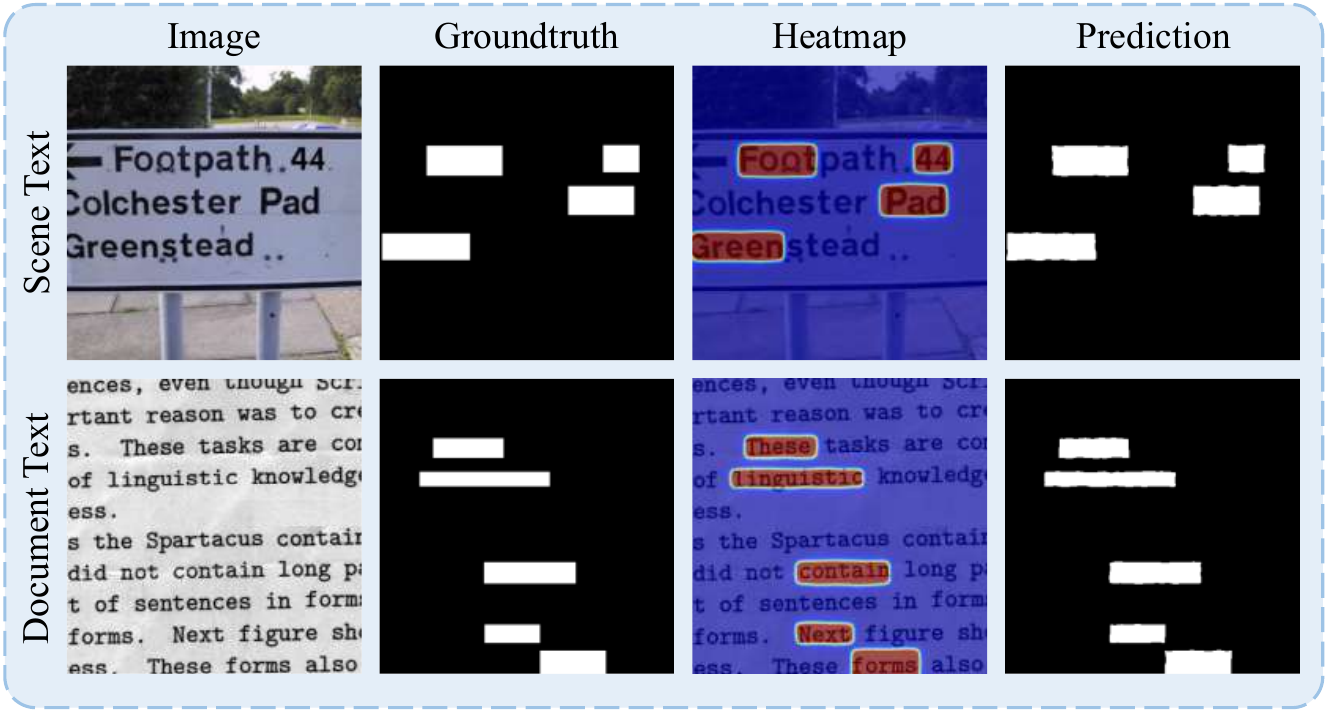}
\caption{Visualization of text tampering localization. Our F-DiT predicts a restoration velocity field (heatmap) whose magnitude indicates the local restoration cost required to reduce deviation from authentic visual-text statistics. The final binary localization mask (right) is obtained by thresholding this velocity magnitude.}
\Description{Two examples of visual text tampering localization are shown in separate rows. The first row contains a natural scene-text
image, and the second row contains a document-text image. From left to right, each row presents the input image, the binary ground-truth tampering mask, the predicted restoration-cost heatmap, and the final binary prediction. Bright regions in the masks and high-response regions in the heatmaps indicate localized tampered text areas.}
\label{fig:framework_overview}
\vspace{-6pt}
\end{figure}

In this work, we shift from discriminating forgery patterns to estimating deviations from authentic visual-text statistics. We model authenticity as an empirical distribution induced by the training data and forensic priors, rather than a universal manifold of all natural images. Under this view, authentic regions are locally consistent with such statistics, while tampered regions introduce subtle inconsistencies. We therefore formulate detection as estimating the \textit{restoration cost} required to reduce the discrepancy between a query image and this distribution, using a vector field whose magnitude indicates tampering. To operationalize this formulation, we adopt Flow Matching (FM) \cite{lipman2023flow} to learn a deviation-aware vector field for local restoration-cost estimation. Our goal is not to fully restore or purify forged images, but to use the predicted field as a detector of local authenticity deviation.

However, directly applying standard FM to forensics faces two challenges. First, tampering is spatially sparse, causing standard FM objectives to be dominated by background regions and leading to a ``lazy'' identity solution in anomalous areas. Second, paired supervision is typically unavailable, requiring the model to learn deviation-aware restoration dynamics in a self-supervised manner. We address these challenges with Sparse-Constraint Rectified Flow (SC-RF). Specifically, we redefine the flow matching objective under a \textit{spatially weighted measure}, which encourages the learned vector field to focus on high-frequency restoration signals in sparse regions. To mitigate the lack of paired supervision, we introduce a self-supervised Artifact Injection training paradigm. Instead of hallucinating content, we train the model to reduce controlled local perturbations, including texture degradation and editing-related local inconsistencies, enabling it to learn a vector field aligned with authentic visual-text statistics rather than performing semantic inpainting.

Furthermore, we design Forensic-DiT, a specialized Diffusion Transformer architecture tailored for this task. Unlike standard DiTs that rely on VAE compression—which irreversibly destroys high-frequency forensic traces—our architecture operates in pixel space and incorporates \textit{physics-aware inductive biases} (e.g., SRM and DCT features) directly into the embedding layer. Extensive experiments show that this formulation yields strong open-set detection performance, particularly in zero-shot settings, while our auxiliary stress-test analysis further suggests that local harmonization can weaken part of the statistical cues used by existing detectors.

% Our contributions can be summarized as follows:
% \begin{itemize}
%     \item We cast open-set visual text forensics as a \emph{generative detection} problem, where tampering is localized by estimating the local restoration cost required to align a query image with \emph{authentic visual-text statistics}. This formulation reduces reliance on forgery-specific decision boundaries and improves generalization to unseen attacks.

%     \item We formulate \emph{Sparse-Constraint Rectified Flow (SC-RF)}, a detector-oriented modification to Flow Matching for spatially sparse anomaly localization. By redefining the transport objective under a spatially weighted measure, SC-RF alleviates the optimization bias toward vast background regions and improves learning on sparse tampered areas.

%     \item We propose \emph{Forensic-DiT}, a physics-aware architecture operating in pixel space to preserve high-frequency forensic traces. Combined with our self-supervised \emph{Artifact Injection} strategy, the model learns a robust restoration-cost field without requiring paired forensic data.

%     \item We achieve state-of-the-art performance on comprehensive benchmarks, particularly in zero-shot settings. In addition, we provide an auxiliary \emph{stress-test analysis} showing that local harmonization can weaken part of the statistical cues used by existing detectors.
% \end{itemize}

Our contributions can be summarized as follows:
\begin{itemize}
    \item We cast open-set visual text forensics as a \emph{generative detection} problem and formulate \emph{Sparse-Constraint Rectified Flow (SC-RF)} to estimate local restoration costs relative to authentic visual-text statistics. By redefining Flow Matching under a spatially weighted measure, SC-RF reduces reliance on forgery-specific decision boundaries and alleviates optimization bias toward vast authentic background regions.

    \item We propose \emph{Forensic-DiT}, a physics-aware pixel-space architecture that preserves high-frequency forensic traces through multimodal RGB, SRM, and DCT representations and image-specific forensic fingerprint modeling. Combined with self-supervised \emph{Artifact Injection}, it learns a robust restoration-cost field from authentic images without requiring paired forensic training data or semantic inpainting supervision.

    \item We achieve state-of-the-art performance across three benchmarks, surpassing the runner-up by 3.2 and 4.8 percentage points in F1 and IoU, respectively, with particularly strong zero-shot generalization to unseen editing patterns. We further provide an auxiliary \emph{stress-test analysis} showing that local harmonization can weaken statistical cues used by existing detectors.
\end{itemize}

\section{Related Work}

\subsection{Image Text Tampering Detection}

Image text tampering detection has evolved from generic image forgery analysis \citep{zhou2018learning} toward more fine-grained forensic modeling with pixel-level localization capabilities \citep{wu2019mantra}. Early deep-learning approaches primarily focused on improving localization accuracy through dense prediction architectures \citep{zhuang2021dfcn}, multi-task learning frameworks \citep{chen2021mv}, and residual feedback mechanisms that enhance sensitivity to subtle manipulation traces \citep{bi2019rrunet}. While these methods achieve strong performance on known manipulation patterns, they largely rely on discriminative cues learned from closed-set training data.

As text manipulation becomes increasingly diverse and visually realistic, generic forgery features gradually lose reliability. To better capture text-specific artifacts, subsequent studies incorporate local texture statistics \citep{cruz2017lbp}, structural decoupling via two-stream architectures \citep{xu2022twostream}, and residual-domain anomaly modeling \citep{BayarS16}. More recent efforts further explore frequency-domain priors \citep{frank2020leveraging,WangXXWZZ22} and character-level feature interactions \citep{liao2023ctpnet, luo2025realdtt}. Nevertheless, these approaches remain predominantly discriminative and often struggle to generalize to unseen or emerging text editing patterns, which reflects a core limitation of closed-set learning \citep{bendale2016towards}. In contrast, we formulate open-set text forensics as generative detection by estimating local deviations from authentic visual-text statistics.

% In contrast, our work treats open-set visual text forensics as a generative detection problem. Rather than learning forgery-specific decision boundaries, we estimate how strongly local regions deviate from authentic visual-text statistics. This detector-oriented formulation is better aligned with open-set scenarios in which manipulation artifacts evolve faster than closed-set supervised labels.

\subsection{Generative Models for Anomaly Detection}

Unsupervised anomaly detection commonly assumes that models trained on normal data fail to accurately reconstruct anomalous samples. Early autoencoder- and GAN-based methods exploit this assumption by using reconstruction degradation as the anomaly signal \citep{sakurada2014anomaly,an2015variational}, including latent optimization in AnoGAN \citep{Schlegl2017fAnoGAN} and memory-constrained reconstruction in MemAE and its variants \citep{Gong2019MemAE}. However, such reconstruction-driven approaches are fundamentally limited when normal data distributions become complex.

Diffusion-based anomaly detection replaces one-shot reconstruction with progressive denoising trajectories, where anomalies persist as residual deviations across steps. AnoDDPM enables anomaly localization through accumulated restoration errors \citep{Wyatt2022AnoDDPM}, while subsequent studies highlight the strong modeling capacity of diffusion priors alongside their reliance on multi-step inference and sensitivity to noise scheduling \citep{Zhang2023DiffusionAD}.

Flow-based generative modeling \citep{rezende2015variational, kingma2018glow} provides a deterministic alternative by formulating generation as continuous probability transport. Flow Matching learns velocity fields that capture structured transport
dynamics between source and target distributions
\citep{lipman2023flow}, while Rectified Flow further
simplifies transport paths for more direct generative modeling \citep{Liu2022RectifiedFlow}. Compared with diffusion models, flow-based methods retain progressive recovery while avoiding heavy iterative sampling, offering a favorable balance between modeling capacity and efficiency.

Despite these advantages, existing generative anomaly detection methods are rarely tailored to text forensics, where anomalies are spatially sparse, highly localized, and often dominated by subtle statistical inconsistencies rather than large semantic deviations. Our method differs from prior reconstruction- or diffusion-based anomaly detectors \citep{Zhang2026IFANet} in three aspects. First, it is designed for detector-oriented localization rather than generic reconstruction. Second, its source distribution is formed by self-supervised local corruption on authentic text images rather than by Gaussian noise or unconstrained anomaly synthesis. Third, it combines flow-based learning with forensic-oriented pixel-space priors, enabling restoration-cost estimation that is better suited to open-set text tampering detection.

\section{Methodology}

\begin{figure*}[t]
  \centering
  \includegraphics[width=\linewidth]{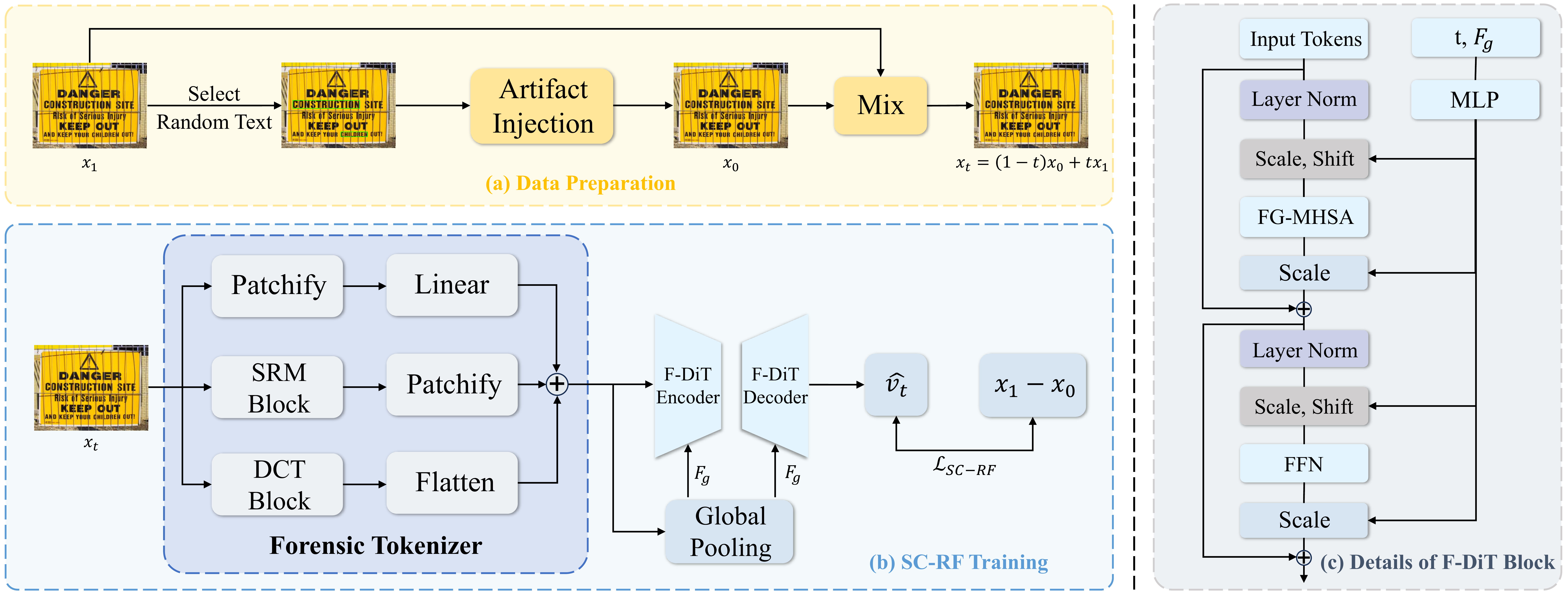}
  \caption{Overview of our proposed framework.
(a) Data Preparation: We synthesize the source distribution $p_0$ via self-supervised Artifact Injection on authentic images.
(b) SC-RF Training: The model is trained to regress the linear restoration trajectory from corrupted $x_0$ to authentic $x_1$ using our Sparse-Constraint Rectified Flow loss.
(c) F-DiT Block: The core architecture features a Multi-modal Forensic Tokenizer (fusing RGB, SRM, DCT features) and injects global forensic fingerprints via SF-AdaLN and Frequency-Gated Multi-Head Self-Attention to capture micro-statistical anomalies.
The model $v_\theta$ predicts a velocity field $\hat{v}_t$ to match the linear transport target $x_1 - x_0$.}
\Description{The framework contains three components. Panel (a) shows data preparation, where a random text region is selected from
an authentic image, locally modified through Artifact Injection, and mixed with the original image to produce a corrupted sample. Panel (b) shows SC-RF training. RGB patches, SRM residual features, and DCT frequency features are processed by the forensic tokenizer and combined before entering the F-DiT encoder and decoder. Global pooling produces a forensic fingerprint, and the network predicts a velocity field that is compared with the restoration target using the SC-RF loss. Panel (c) shows an F-DiT block composed of layer normalization, scale-and-shift modulation, frequency-gated multi-head self-attention, a feed-forward network, and residual connections.}
  \label{fig:framework}

\end{figure*}

\subsection{Preliminaries}

Flow Matching (FM) \cite{lipman2023flow, AlbergoV23} models the probability density path evolving from a source distribution $p_0$ to a target distribution $p_1$ via a time-dependent vector field $v_t$. This evolution is governed by the ODE $d x_t / d t = v_t(x_t)$. In this work, we adopt the \textit{Rectified Flow} \cite{Liu2023RectifiedFlow} formulation, which constructs an Optimal Transport (OT) displacement path between a sample pair $(x_0, x_1)$. The interpolation follows a straight line geodesic:
\begin{equation}
x_t = (1 - t)x_0 + t x_1, \quad t \in [0, 1]
\label{eq:interpolation}
\end{equation}
where $x_0 \sim p_0$ and $x_1 \sim p_1$. The ground-truth conditional vector field $u_t$ thus takes a simple, closed form:
\begin{equation}
u_t(x_t|x_0, x_1) = x_1 - x_0
\end{equation}
The model is trained to regress this velocity field by minimizing the following objective:
\begin{equation}
\mathcal{L}_{\text{RF}}(\theta) = \mathbb{E}_{t, x_0, x_1} \left[ \left\| v_\theta(x_t, t) - (x_1 - x_0) \right\|^2 \right]
\label{eq:cfm_loss}
\end{equation}
% While standard generative modeling assumes $p_0$ is Gaussian noise, we redefine the boundary conditions to model the transition from a corrupted manifold to an authentic manifold for tampering detection.
While standard generative modeling assumes $p_0$ is Gaussian noise, we redefine the boundary conditions to model the transition from a locally corrupted distribution to an authentic visual-text distribution for tampering detection.

\subsection{Sparse-Constraint Rectified Flow (SC-RF)}

While standard Rectified Flow is effective for probability transport, directly applying it to open-set tampering detection presents unique challenges. We propose Sparse-Constraint Rectified Flow (SC-RF), a detector-oriented adaptation of Flow Matching for spatially sparse, high-frequency anomalies.

\textbf{Redefining the Flow Boundaries.}
In our forensic setting, the target distribution $p_1(x)$ represents the empirical distribution of \textit{authentic visual-text statistics}, denoted by $\mathcal{M}_r$. Importantly, $\mathcal{M}_r$ is not intended to represent a universal manifold of all natural images, but rather the authentic distribution induced by our training corpus and forensic priors. We redefine the source distribution $p_0(x)$ not as Gaussian noise, but as a \textit{locally corrupted distribution} $\mathcal{M}_c$, where samples $x_0 \sim p_0(x)$ are synthesized from authentic images and contain local texture anomalies (e.g., interpolation inconsistencies, compression mismatch, or noise inconsistency). The flow trajectory $x_t$ therefore models a detector-oriented restoration path from a locally corrupted observation toward authentic visual-text statistics.

\textbf{The Sparsity Challenge.}
A critical issue arises from the spatial nature of tampering. For a given pair $(x_0, x_1)$, the tampering mask $M \in \{0, 1\}^{H \times W}$ indicates modified regions. The relationship is given by:
\begin{equation}
x_0 = x_1 \odot (1 - M) + \mathcal{T}(x_1) \odot M
\end{equation}
where $\mathcal{T}$ is the tampering operator. The target vector field $u = x_1 - x_0$ is consequently \textit{sparse}:
\begin{equation}
u(s) = \begin{cases}
x_1(s) - \mathcal{T}(x_1(s)) \neq 0, & \text{if } M(s) = 1 \\
0, & \text{if } M(s) = 0
\end{cases}
\end{equation}
In practice, tampered regions often occupy only a negligible fraction of the image, often less than $5\%$. If we directly minimize the standard RF loss (Eq.~\ref{eq:cfm_loss}), the optimization is dominated by the vast background regions where the trivial solution $v_\theta \approx 0$ yields near-zero loss. As a result, the gradients associated with sparse anomalous regions are severely under-emphasized, causing the model to degenerate toward an identity mapping on the very regions of interest.

\textbf{Sparse-Weighted Optimal Transport.}
To address this issue, we reformulate the flow matching objective by altering the integration measure over the spatial domain $\Omega$. Instead of the uniform Lebesgue measure, we introduce a \textit{Sparsity-Weighted Measure} $\mu_w$ parameterized by the tampering mask:
\begin{equation}
d\mu_w(s) = (1 + \lambda \cdot M(s)) ds
\end{equation}
where $\lambda \gg 1$ is a hyperparameter controlling the penalty on sparse anomalous regions. Minimizing the transport cost under this non-uniform measure encourages the learned vector field to fit the restoration target more faithfully within sparse tampered areas while maintaining stability in the background. Our proposed \textbf{SC-RF Objective} is:
% \begin{multline}
% \mathcal{L}_{\text{SC-RF}}(\theta)
% = \mathbb{E}_{t, x_0, x_1} \Bigg[
% \frac{1}{|\Omega|}
% \sum_{s \in \Omega}
% \mathbf{W}_s
% \big\|
% v_\theta(x_t, t)_s \\
% - (x_1 - x_0)_s
% \big\|^2
% \Bigg]
% \label{eq:scrf_loss}
% \end{multline}

\begin{equation}
\mathcal{L}_{\mathrm{SC-RF}}(\theta)
=
\mathbb{E}_{t,x_0,x_1}\!\left[
\frac{1}{|\Omega|}
\sum_{s\in\Omega}
\mathbf{W}_s
\left\|v_\theta(x_t,t)_s-u_s\right\|_2^2
\right]
\label{eq:scrf_loss}
\end{equation}

where the spatial weight map is defined as $\mathbf{W}_s = 1 + \lambda \cdot M_s$.
Here, $\Omega$ denotes the lattice of pixel positions (e.g., $H \times W$), with $|\Omega|$ being the total number of pixels.
The symbol $s \in \Omega$ denotes a spatial pixel index rather than a probability variable.
Moreover, $v_\theta(x_t,t)$ denotes the predicted velocity field, while $u_s=(x_1-x_0)_s$ denotes the ground-truth transport target at location $s$.
% Moreover, $v_\theta(x_t, t)$ (denoted as $\hat{v}_t$ in Fig.~2) represents the predicted velocity field, while $x_1 - x_0$ serves as the ground-truth transport target.

\textbf{Detector-Oriented Interpretation.}
This objective can be interpreted as learning a \textit{restoration-cost field}. By imposing a heavy penalty $\lambda$ on tampered regions, we encourage the network to predict a high-magnitude vector specifically at locations where local texture statistics deviate from the authentic distribution $\mathcal{M}_r$. During inference, the magnitude of this predicted vector, $\|v_\theta(x, 0)\|$, serves as a direct measure of the \textit{local restoration cost} required to reduce the discrepancy between the query image and authentic visual-text statistics. We do not interpret this field as a guarantee of full image restoration or exact projection onto a universal real-image manifold; rather, it is used as a detector-oriented anomaly score for local tampering localization.

\subsection{Forensic-DiT Architecture}

To estimate the restoration-cost field $v_t$, we propose F-DiT. We adopt U-DiT~\cite{Tian2024UDiTs} as our backbone, which combines the long-range dependency modeling of Transformers~\cite{vaswani2017attention} with U-Net-style skip connections~\cite{ronneberger2015u}, making it well suited for pixel-level dense prediction. Unless otherwise specified, the scale/shift operators in Fig.~\ref{fig:framework} follow the residual modulation design of the U-DiT backbone and are not introduced as additional standalone modules. On top of this backbone, we introduce several forensic-oriented modifications to better capture microscopic tampering traces and authenticity-related local statistics.

\textbf{Multi-modal Forensic Tokenizer.}
Standard patchification often discards subtle high-frequency artifacts. We propose a hybrid tokenizer fusing three domains. Given an input $x_t$, we extract: (1) Visual tokens from RGB patches; (2) Noise tokens from Spatial Rich Model (SRM) residuals~\cite{FridrichK12}; and (3) Frequency tokens from flattened Block-DCT coefficients. These features are projected to a shared latent dimension $d$ via embeddings $E(\cdot)$ and fused via element-wise summation:
\begin{equation}
h = E_v(x_t) + E_n(\mathcal{F}_s(x_t)) + E_f(\mathcal{F}_d(x_t))
\end{equation}
where $\mathcal{F}_s$ and $\mathcal{F}_d$ denote SRM filtering and Block-DCT operations, respectively. This ensures each token $h$ encodes both visual content and local signal-to-noise characteristics.

\textbf{Self-Fingerprint Adaptive Normalization (SF-AdaLN).}
Rather than relying on a fixed dataset-level notion of authenticity, we use the dominant forensic statistics of each image as an image-specific reference. A Global Forensic Fingerprint $F_g$ is computed by global average pooling over the input tokens $h$. We replace the standard adaLN with SF-AdaLN, where the scale $\gamma$ and shift $\beta$ parameters are regressed from both the time embedding $e_t$ and the global fingerprint:
\begin{equation}
(\gamma, \beta) = \text{MLP}(e_t + \mathcal{P}(F_g))
\end{equation}
where $\mathcal{P}$ denotes a linear projection layer. This design allows the detector to measure local deviations relative to the \emph{image-specific forensic background}, which is important because authentic scene text, scanned documents, and photographed documents may exhibit substantially different noise, compression, and acquisition statistics. As a result, SF-AdaLN improves robustness to benign domain variation while making anomalous local inconsistencies easier to identify.

\textbf{Frequency-Gated Multi-Head Self-Attention.}
Standard attention calculates semantic similarity, which may erroneously correlate tampered regions with the background due to semantic blending. To address this, we introduce a Frequency Gate to sever connections between tokens with distinct spectral statistics. Specifically, we define a Frequency Bias matrix $\mathbf{B}$ derived from the frequency features $h_f$ (extracted by $E_f$). For tokens $i$ and $j$, the bias is computed as:
\begin{equation}
    \mathbf{B}_{i,j} = \text{MLP}(|h_f^{(i)} - h_f^{(j)}|)
\end{equation}
This term measures the forensic discrepancy between two patches. The Frequency-Gated Multi-Head Self-Attention is then defined as:
\begin{equation}
    \text{Attention}(Q, K, V) = \text{Softmax}\left( \frac{QK^T}{\sqrt{d}} - \eta \cdot \mathbf{B} \right) V
\end{equation}

% where $\eta$ is a learnable scalar. The subtractive form treats $\mathbf{B}$ as a discrepancy penalty in the attention-logit space: patch pairs with larger spectral inconsistency receive smaller attention weights. Compared with multiplicative gating, the subtractive form is more stable because it suppresses discrepancy directly in the logit space without entangling semantic similarity and forensic inconsistency. Additive modulation with a negative bias is functionally related, but the subtractive form is more explicit and interpretable. Consequently, this mechanism forces the model to aggregate context only from regions with similar texture statistics, preventing tampered regions from ``stealing'' authentic features to camouflage themselves.

where $\eta$ is a learnable scalar. The negative bias suppresses attention between patches with inconsistent spectral statistics, encouraging context aggregation among forensically similar regions.

\subsection{Self-Supervised Training}
\label{sec:self-supervised_training}
\textbf{Artifact Injection.}
To mitigate data scarcity and enhance open-set generalization, we propose Artifact Injection to synthesize locally corrupted samples $x_0$ from authentic images $x_1$. We randomly select text regions and inject localized micro-artifacts, forcing the model to learn tampering detection through \emph{local inconsistency identification} rather than forgery-specific supervision. The injection applies a random combination of five operations: Gaussian blur, JPEG compression, Gaussian noise, local alpha-blending boundary mismatch, and local glyph re-rendering mismatch. 

Specifically, local alpha-blending boundary mismatch pastes a text patch into the selected region with controlled alpha blending and fixed boundary feathering, which introduces transition inconsistency between the edited patch and the surrounding background. Local glyph re-rendering mismatch replaces the selected text region with a newly rendered glyph patch under controlled font rasterization, stroke sharpness, color quantization, and local compression settings, which creates appearance inconsistency between the edited text and the original image statistics.

% All operations are applied locally rather than globally so that the model learns to detect inconsistent local statistics instead of uniformly degraded image quality. These synthetic perturbations are not designed to exactly replicate every real editing pipeline. Their purpose is to expose the detector to a broad and diverse set of authenticity-related local deviations that frequently appear in practical text editing scenarios. The selection of operations, their intensities, and their composition order are all randomized to maximize source-distribution diversity and reduce overfitting to any single perturbation pattern.

All operations are applied locally, with randomized types, intensities, and orders, to expose the detector to diverse local inconsistencies rather than replicate specific editing pipelines.

\textbf{Curriculum Training.}
Directly training on highly sparse anomalies can be unstable. We therefore design a Three-Stage Curriculum based on the sparsity ratio of injected artifacts. In Stage 1 (Warm-up), artifacts are injected into a large proportion (e.g., 50\%--80\%) of text regions to provide dense supervision signals. In Stage 2 (Transition), we linearly decrease the injection ratio. Finally, in Stage 3 (Refinement), we train on realistic sparse settings (e.g., $<5\%$) using the proposed Sparse-Constraint Rectified Flow loss. This easy-to-hard schedule allows the model to first learn \textit{how} to estimate restoration targets and then adapt to \textit{where} anomalous local corrections should be emphasized in highly sparse scenarios.

\subsection{Inference as Local Restoration-Cost Estimation}
\label{sec:inference}

% At inference time, we do not integrate the full ODE trajectory. Instead, given a query image $x_q$, we treat it as the input state at $t=0$ and feed it into F-DiT to predict the instantaneous velocity field $\hat{v}_0 = v_\theta(x_q, 0)$. Here, $\hat{v}_0$ is not interpreted as a recovered image or a complete restoration path. Rather, it is interpreted as an instantaneous local correction field that indicates how strongly each spatial location needs to be adjusted to better match authentic visual-text statistics.

At inference, we avoid ODE integration and directly predict the instantaneous velocity field $\hat{v}_0=v_\theta(x_q,0)$, whose magnitude measures the local correction required to match authentic visual-text statistics. For a spatial location $s$, the predicted vector $\hat{v}_{0,s}$ contains channel-wise correction values. Authentic regions typically require little correction and therefore yield $\|\hat{v}_{0,s}\|_2 \approx 0$, whereas tampered regions usually require larger corrections and thus produce higher-magnitude vectors. We therefore convert the predicted vector field into a scalar tampering score map by taking the $L_2$ norm across channels:
\begin{equation}
\hat{M}_s = \|\hat{v}_{0,s}\|_2,
\end{equation}
where $s$ denotes the spatial location. The resulting map $\hat{M}$ is used as the pixel-level tampering probability map. In this way, the model uses the magnitude of the predicted instantaneous correction, rather than an explicitly integrated restoration trajectory, as the anomaly score for tampering localization.

\section{Experiments}
\subsection{Experimental Setup}

\textbf{Datasets.} To construct a diverse empirical distribution of authentic visual-text statistics, we compile a self-supervised training set spanning both natural scenes and documents. Following the protocol established in DAF \citep{Qu2025RevisitingTampered} , we utilize a collection of scene text datasets: LSVT \citep{sun2019lsvt}, ReCTS \citep{zhang2019rects}, ICDAR2013 \citep{karatzas2013rr}, ICDAR2015 \citep{karatzas2015rr}, ICDAR2017 \citep{nayef2017mlt}, TextOCR \citep{singh2021textocr}, and ArT \citep{chng2019art}. Crucially, to extend the model's robustness to document-specific textures and geometric distortions, we further incorporate RVL-CDIP \citep{harley2015rvlcdip} into the training corpus. For evaluation, we employ three complementary benchmarks to ensure a holistic assessment: the large-scale document image dataset DocTamper \citep{QuLLCPGJ23}, the classic scene text tampering benchmark Tampered-IC13 (T-IC13) \citep{WangXXWZZ22}, and the recently proposed OSTF \citep{Qu2025RevisitingTampered}. Specifically, OSTF serves as a crucial testbed for open-set generalization, featuring forgeries generated by various state-of-the-art text editing methods under cross-source settings.

\begin{table*}[t]

\centering

\caption{Quantitative comparison with state-of-the-art methods. Pixel-level F1, IoU, and AUC scores are evaluated across three benchmarks: T-IC13, DocTamper, and OSTF. ``Avg'' denotes the mean over the three benchmarks. ``z.'' and ``f.'' denote zero-shot and full-shot settings, respectively. \textbf{Bold} and \underline{underlined} values indicate the best and second-best results, respectively.}

\label{tab:main_results}

\setlength{\tabcolsep}{6pt} % column padding

\renewcommand{\arraystretch}{1.05} % row height

\small

\begin{tabular}{c ccc ccc ccc ccc}

\toprule

\multirow{2}{*}{\centering \textbf{Method}}

& \multicolumn{3}{c}{\textbf{T-IC13}}

& \multicolumn{3}{c}{\textbf{DocTamper}}

& \multicolumn{3}{c}{\textbf{OSTF} }

& \multicolumn{3}{c}{\textbf{Avg}}\\

\cmidrule(lr){2-4} \cmidrule(lr){5-7} \cmidrule(lr){8-10}\cmidrule(lr){11-13} 

& F1 & IoU & AUC & F1 & IoU & AUC & F1 & IoU & AUC & F1 & IoU & AUC \\

\midrule

MVSS-Net (f.)       & 0.553 & 0.382 & 0.647 & 0.431 & 0.275 & 0.683 & 0.353 & 0.214 & 0.556 & 0.446 & 0.290 & 0.629\\

PSCC-Net (f.)       & 0.504 & 0.337 & 0.594 & 0.384 & 0.238 & 0.656 & 0.314 & 0.186 & 0.525 & 0.401 & 0.254 & 0.592\\

DeepLabV3+ (f.)     & 0.744 & 0.592 & 0.895 & 0.651 & 0.483 & 0.809 & 0.495 & 0.329 & 0.661 & 0.630 & 0.468 & 0.788\\

SegFormer (f.)      & 0.790 & 0.653 & 0.928 & 0.604 & 0.433 & 0.783 & 0.521 & 0.352 & 0.682 & 0.638 & 0.479 & 0.798\\

DTD (f.)            & 0.907 & 0.830 & 0.945 & 0.792 & 0.656 & 0.895 & 0.625 & 0.455 & 0.767 & 0.775 & 0.647 & 0.869\\

DAF (f.)            & 0.925 & 0.860 & \textbf{0.973} & 0.759 & 0.612 & 0.851 & 0.759 & 0.612 & 0.884 & 0.814 & 0.695 & 0.903\\

TTDMamba (f.)       & \underline{0.929} & \underline{0.867} & 0.917 & \underline{0.806} & \underline{0.675} & \textbf{0.917} & 0.775 & 0.633 & 0.896 & \underline{0.837} & \underline{0.725} & \underline{0.910}\\

Ours \textbf{(z.)}  & 0.905 & 0.826 & 0.939 & 0.741 & 0.589 & 0.822 & \underline{0.781} & \underline{0.641} & \underline{0.904} & 0.809 & 0.685 & 0.888\\

Ours (f.)           & \textbf{0.941} & \textbf{0.889} & \underline{0.961} & \textbf{0.845} & \textbf{0.732} & \underline{0.913} & \textbf{0.822} & \textbf{0.698} & \textbf{0.925} & \textbf{0.869} & \textbf{0.773} & \textbf{0.933}\\

\bottomrule

\end{tabular}

\end{table*}

\textbf{Evaluation Metrics.} Following established protocols in image manipulation detection \citep{QuLLCPGJ23, duan2025realdtt}
, we formulate the text tampering detection task as a pixel-level binary classification problem. To provide a comprehensive assessment of localization accuracy, we adopt the Intersection over Union (IoU) and F1-score as our primary metrics. Furthermore, to evaluate the model's discriminative robustness independent of specific decision thresholds, we additionally report the Area Under the Receiver Operating Characteristic Curve (AUC).

\textbf{Baselines.}
We compare our method with a diverse set of representative baselines spanning different methodological paradigms.
Specifically, we include two generic semantic segmentation models, \emph{DeepLabV3+} \citep{chen2018deeplabv3plus} and \emph{SegFormer} \citep{xie2021segformer},
to establish discriminative segmentation baselines without explicit modeling of tampering cues.
We further consider general image manipulation localization approaches, including \emph{MVSS-Net} \citep{chen2021mv} and \emph{PSCC-Net} \citep{liu2022pscc},
which leverage forensic-oriented visual inconsistencies for detecting manipulated regions.
In addition, we evaluate against text tampering--specific methods, including \emph{DTD} \citep{QuLLCPGJ23},
\emph{DAF} \citep{Qu2025RevisitingTampered}, and \emph{RealDTT} \citep{duan2025realdtt} with its corresponding model \emph{TTDMamba},
which represent recent advances in document- and scene-level tampered text detection under both synthetic and real-world settings.
This setting enables a fair comparison across discriminative, forensic-driven, and text-specific detection paradigms.
For methods that were not originally evaluated on our selected benchmarks, we re-train them using the same training sets,
adopting the training strategy and parameter configurations recommended in the original papers to ensure fair comparison.

\begin{figure*}[!t]
\centering
\includegraphics[width=\linewidth]{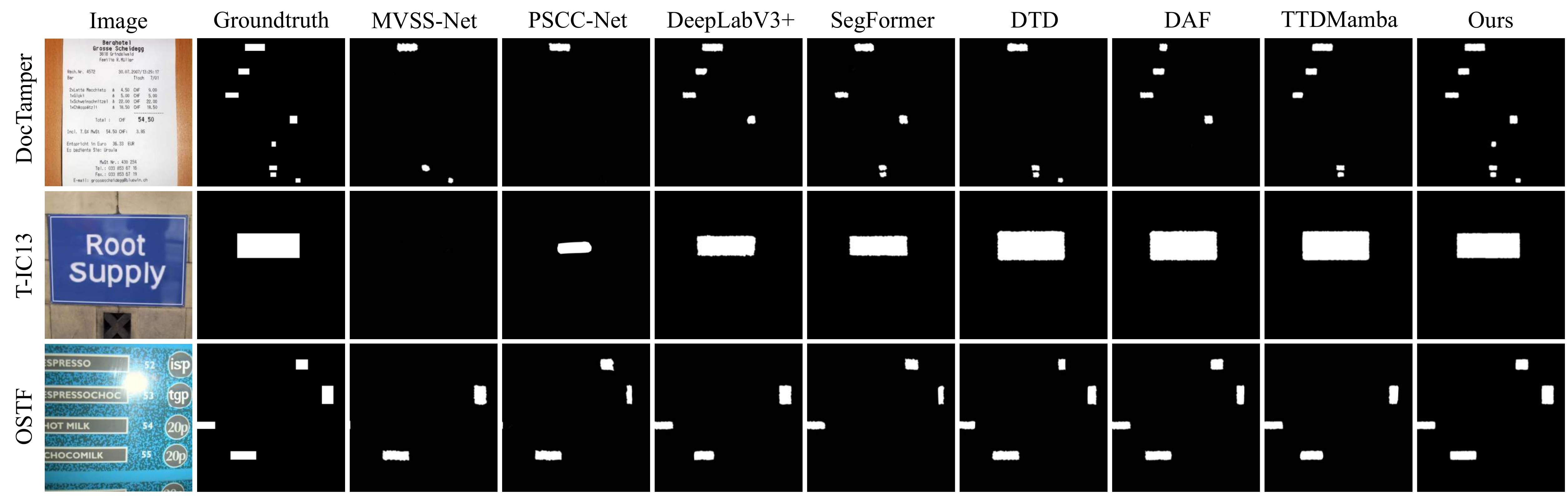}
\caption{Qualitative comparison of tampering localization results on challenging samples from OSTF, DocTamper, and T-IC13 benchmarks.}
\Description{Three rows compare text tampering localization results on examples from DocTamper, T-IC13, and OSTF. Each row shows the
input image, the binary ground-truth mask, and the masks predicted by MVSS-Net, PSCC-Net, DeepLabV3+, SegFormer, DTD, DAF, TTDMamba,
and the proposed method. Tampered regions are displayed in white against a black background. The predictions vary in the number,
extent, and boundary accuracy of the detected regions, while the proposed method produces masks that visually align closely with the ground-truth tampered areas.}
\label{fig:qualitative_comparison}
\end{figure*}

\textbf{Implementation Details.} Our proposed framework is implemented in PyTorch and trained on a workstation with four NVIDIA RTX A6000 GPUs. Distinct from latent diffusion models, we operate directly in pixel space without using Variational Autoencoders (VAE) for downsampling, so as to preserve high-frequency forensic artifacts. Input images are resized to $512 \times 512$, and the patch size is set to 4 to balance efficiency and fine-grained feature extraction. The model is optimized with AdamW using an initial learning rate of $1 \times 10^{-4}$, a weight decay of $1 \times 10^{-4}$, cosine annealing, and a linear warmup of 5,000 steps. The global batch size is 16. Training runs for 300k iterations, and Exponential Moving Average (EMA) with decay 0.9999 is applied at inference. Unless otherwise specified, the sparsity weight in Eq.~(6) is set to $\lambda=20$, and the frequency-gating scalar in Eq.~(11) is initialized as $\eta=1.0$ and optimized jointly with the network. The three-stage curriculum uses 60k, 90k, and 150k iterations for warm-up, transition, and refinement, respectively; the artifact coverage ratio is sampled from 50\%--80\% in Stage 1, decreased linearly from 50\% to 5\% in Stage 2, and sampled from 1\%--5\% in Stage 3. For evaluation, F1 and IoU are computed by thresholding the predicted tampering score map with a single validation-selected threshold fixed for all test images within each benchmark, using $\tau=0.34$ for T-IC13, $\tau=0.31$ for DocTamper, and $\tau=0.33$ for OSTF, while AUC is computed directly from the raw score map.

\begin{table*}[t]
\centering
% \caption{Adversarial harmonization analysis. The degradation in detection performance(F1) of existing forensic models is evaluated when input images are processed by the proposed F-DiT. ``Ori.'' and ``Har.'' denote results before and after F-DiT harmonization, respectively. ``Avg'' denotes the mean over the three benchmarks. \textbf{Bold} values indicate the largest performance degradation among the evaluated forensic models for each benchmark and the Avg column.}
\caption{Stress-test analysis under local harmonization. The degradation in detection performance (F1) of existing forensic models is evaluated when input images are processed by the proposed F-DiT. ``Ori.'' and ``Har.'' denote results before and after F-DiT processing, respectively. ``Avg'' denotes the mean over the three benchmarks. Bold values indicate the largest performance degradation among the evaluated forensic models for each benchmark and the Avg column.}
\label{tab:adversarial-harmonization-analysis}
\setlength{\tabcolsep}{6pt}
\renewcommand{\arraystretch}{1.05}

\small
\begin{tabular}{c ccc ccc ccc ccc}
\toprule
\multirow{2}{*}{\textbf{Method}}
& \multicolumn{3}{c}{\textbf{T-IC13}}
& \multicolumn{3}{c}{\textbf{DocTamper}}
& \multicolumn{3}{c}{\textbf{OSTF}}
& \multicolumn{3}{c}{\textbf{Avg} }\\
\cmidrule(lr){2-4} \cmidrule(lr){5-7} \cmidrule(lr){8-10} \cmidrule(lr){11-13}
& Ori. & Har. & $\Delta$
& Ori. & Har. & $\Delta$
& Ori. & Har. & $\Delta$
& Ori. & Har. & $\Delta$ \\
\midrule

MVSS-Net   & 0.553 & 0.382 & $\downarrow 0.171$
           & 0.431 & 0.298 & $\downarrow 0.133$
           & 0.353 & 0.245 & $\downarrow 0.108$
           & 0.446 & 0.308 & $\downarrow 0.137$ \\

PSCC-Net   & 0.504 & 0.315 & $\downarrow 0.189$
           & 0.384 & 0.256 & $\downarrow 0.128$
           & 0.314 & 0.201 & $\downarrow 0.113$
           & 0.401 & 0.257 & $\downarrow 0.143$ \\

DeepLabV3+ & 0.744 & 0.528 & $\downarrow 0.216$
           & 0.651 & 0.485 & $\downarrow 0.166$
           & 0.495 & 0.362 & $\downarrow 0.133$
           & 0.630 & 0.458 & $\downarrow 0.172$ \\

SegFormer  & 0.790 & 0.565 & $\downarrow 0.225$
           & 0.604 & 0.442 & $\downarrow 0.162$
           & 0.521 & 0.385 & $\downarrow 0.136$
           & 0.638 & 0.464 & $\downarrow 0.174$ \\

DTD        & 0.907 & 0.582 & $\boldsymbol{\downarrow 0.325}$
           & 0.792 & 0.544 & $\boldsymbol{\downarrow 0.248}$
           & 0.625 & 0.418 & $\downarrow 0.207$
           & 0.775 & 0.515 & $\downarrow 0.260$ \\

DAF        & 0.925 & 0.612 & $\downarrow 0.313$
           & 0.759 & 0.528 & $\downarrow 0.231$
           & 0.759 & 0.503 & $\boldsymbol{\downarrow 0.256}$
           & 0.814 & 0.548 & $\boldsymbol{\downarrow 0.267}$ \\

TTDMamba   & 0.929 & 0.715 & $\downarrow 0.214$
           & 0.806 & 0.622 & $\downarrow 0.184$
           & 0.775 & 0.589 & $\downarrow 0.186$
           & 0.837 & 0.642 & $\downarrow 0.195$ \\

\bottomrule
\end{tabular}
\end{table*}

\subsection{Open-Set Detection Performance}
We conduct a comprehensive comparison with representative pixel-level text tampering localization methods, including generic manipulation detectors (MVSS-Net, PSCC-Net), semantic segmentation backbones (DeepLabV3+, SegFormer), and recent text-specific approaches (DTD, DAF, and TTDMamba). Evaluations are performed under both zero-shot (z.) and full-shot (f.) settings on Tampered-IC13, DocTamper, and OSTF. Here, zero-shot denotes direct evaluation after self-supervised training on authentic corpora without benchmark-specific supervised fine-tuning, while full-shot denotes further supervised fine-tuning on the target benchmark. Quantitative results measured by F1-score, IoU, and AUC are reported in Table~\ref{tab:main_results}, with qualitative comparisons shown in Figure~\ref{fig:qualitative_comparison}. These experiments primarily evaluate the effectiveness of our method as an open-set tampering detector.

As shown in Table~\ref{tab:main_results}, our method achieves the best average F1, IoU, and AUC performance across the three benchmarks. In the zero-shot setting, it consistently outperforms generic manipulation detectors and segmentation-based baselines, while remaining competitive with or superior to several fully supervised text-specific methods. Notably, on the challenging OSTF benchmark with diverse unseen text editing patterns, our zero-shot model demonstrates strong localization accuracy, indicating that the proposed detector generalizes well beyond the specific tampering patterns observed during training.

With full-shot fine-tuning, our method further improves and attains the highest average performance on Tampered-IC13, DocTamper, and OSTF. The consistent gains from zero-shot to full-shot settings suggest that the learned representations provide a strong detector initialization while avoiding excessive reliance on dataset-specific artifacts.

Qualitative results in Figure~\ref{fig:qualitative_comparison} further corroborate these findings. Generic manipulation detectors often miss spatially sparse tampered regions, while semantic segmentation models tend to produce fragmented or over-activated predictions. In contrast, our method yields more compact and boundary-consistent localization across both document images and natural scenes.

\subsection{Stress-Test Analysis}
\textbf{Local Harmonization as an Auxiliary Stress Test.}
We treat this experiment as an auxiliary stress test rather than the primary goal of our method. Specifically, we examine whether local harmonization produced by F-DiT can weaken the cues relied upon by existing detectors. As shown in Table~\ref{tab:adversarial-harmonization-analysis}, when input images are processed by F-DiT, the detection performance of all seven baseline methods degrades significantly across three benchmarks. Notably, text-specific forensic methods experience the sharpest declines, with DTD and DAF suffering average F1-score drops of 0.260 and 0.267, respectively. Even the robust TTDMamba exhibits a 0.195 drop. These results suggest that a substantial fraction of the cues used by existing detectors can be weakened by local statistical harmonization.

\begin{figure}[t]
    \centering
    \includegraphics[width=\linewidth]{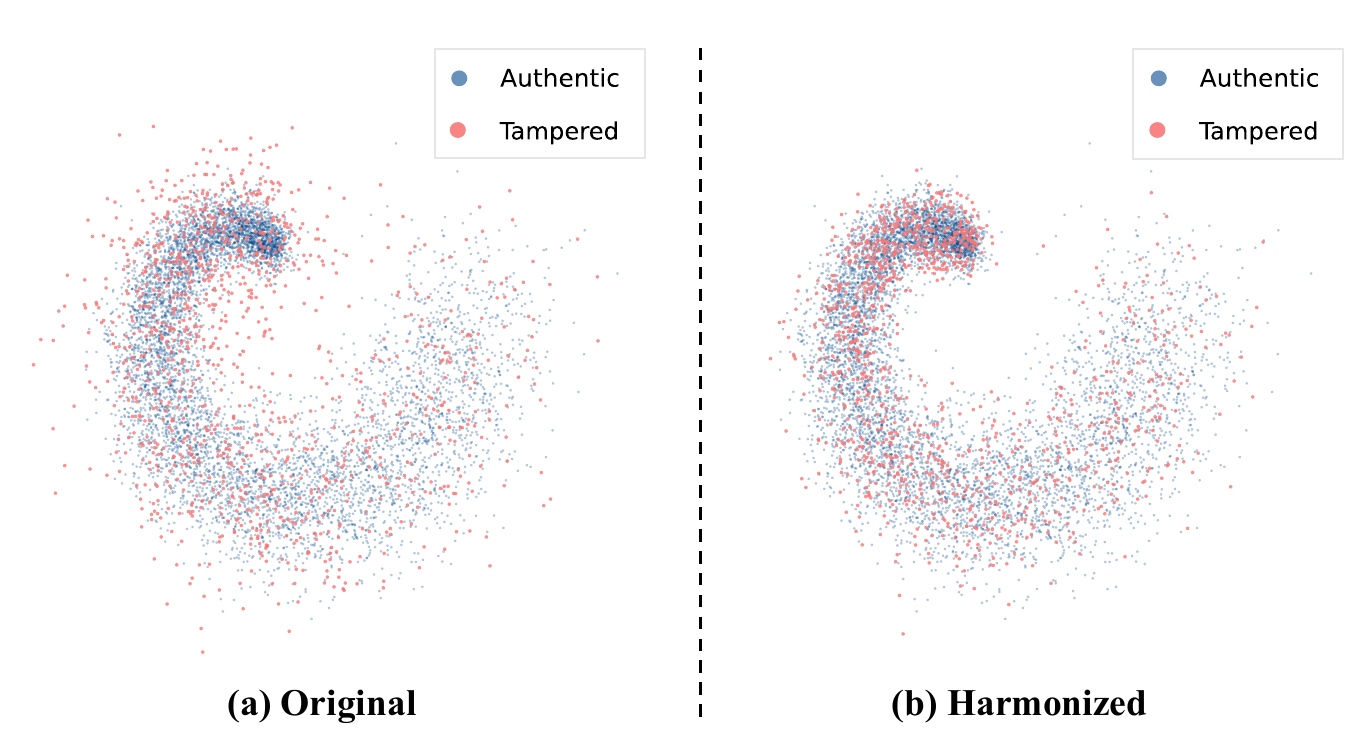} 
    % \caption{t-SNE visualization of feature distributions. (a) In the original space, tampered samples (red) deviate from the authentic manifold (blue). (b) After harmonization by our F-DiT, tampered features are pulled towards the authentic distribution, validating our manifold projection hypothesis.}
    \caption{t-SNE visualization of feature distributions. (a) In the original feature space, tampered samples (red) are separated from authentic samples (blue). (b) After processing by F-DiT, tampered samples shift toward the authentic cluster in the learned feature space.}
    \Description{Two side-by-side t-SNE scatter plots compare the feature distributions of authentic and tampered samples. Blue
    points represent authentic samples, and red points represent tampered samples. In panel (a), labeled Original, the red and blue
    samples form visibly separated distributions. In panel (b), labeled Harmonized, the red samples move toward and overlap more strongly with the blue authentic-sample cluster.}
    \label{fig:tsne_visualization}
\end{figure}

\textbf{Is it Harmonization or Perturbation?}
A critical question is whether the observed performance drop stems from meaningful local harmonization or merely from random perturbation that pushes samples into out-of-distribution regions. To examine the nature of this shift, we visualize the feature-space distribution using t-SNE in Figure~\ref{fig:tsne_visualization}. In the original feature space (Figure~\ref{fig:tsne_visualization}(a)), authentic and tampered samples form two distinct, separable clusters, confirming the existence of a distributional gap. After processing by F-DiT (Figure~\ref{fig:tsne_visualization}(b)), the tampered samples do not scatter randomly; instead, they shift toward the authentic cluster in the learned feature space. This directional movement suggests that F-DiT is not merely adding unstructured perturbations, but is partially reducing the statistical discrepancy exploited by existing detectors. This finding provides a plausible explanation for the results in Table~\ref{tab:adversarial-harmonization-analysis}: after processing by F-DiT, the statistical discrepancy exploited by existing detectors appears to be reduced in the learned feature space, which makes their decision boundaries less effective. We interpret this as supporting evidence for the stress-test analysis, rather than as proof of complete 
restoration or exact projection onto a universal authentic manifold.
\subsection{Ablation Studies}
We validate the proposed design through three complementary ablation studies. Table~\ref{tab:ablation} examines the impact of pixel-space modeling and the forensic-oriented architectural components. Table~\ref{tab:fm_vs_direct_regression} compares Flow Matching with a direct regression objective under the same backbone and training pipeline. Table~\ref{tab:inference_steps_ablation} studies the effect of the number of inference steps in the zero-shot setting.

\begin{table}[t]
\centering
\caption{Zero-shot ablation study on different components. ``VAE'' denotes Variational Autoencoder compression, ``MFT'' denotes Multi-modal Forensic Tokenizer, ``SFA'' denotes Self-Fingerprint Adaptive Normalization, and ``FG-MHSA'' denotes Frequency-Gated Self-Attention with an explicit frequency bias. ``Avg'' denotes the mean over the three benchmarks.}
\label{tab:ablation}
\setlength{\tabcolsep}{2.0pt}
\renewcommand{\arraystretch}{1.05}
\small
\begin{tabular}{l cc cc cc cc}
\toprule
\multirow{2}{*}{\textbf{Method}} 
& \multicolumn{2}{c}{\textbf{T-IC13}} 
& \multicolumn{2}{c}{\textbf{DocTamper}} 
& \multicolumn{2}{c}{\textbf{OSTF}}
& \multicolumn{2}{c}{\textbf{Avg}} \\
\cmidrule(lr){2-3} \cmidrule(lr){4-5} \cmidrule(lr){6-7} \cmidrule(lr){8-9}
& F1 & IoU & F1 & IoU & F1 & IoU & F1 & IoU \\
\midrule

Baseline
& 0.516 & 0.348
& 0.303 & 0.179
& 0.343 & 0.207
& 0.387 & 0.244 \\

w/ VAE
& 0.414 & 0.261
& 0.252 & 0.144
& 0.423 & 0.268
& 0.363 & 0.224 \\

w/o MFT
& 0.827 & 0.705
& 0.690 & 0.527
& 0.685 & 0.521
& 0.734 & 0.584 \\

w/o SFA
& 0.892 & 0.805
& 0.713 & 0.554
& 0.744 & 0.592
& 0.783 & 0.650 \\

w/o FG-MHSA
& 0.866 & 0.764
& 0.705 & 0.544
& 0.743 & 0.591
& 0.771 & 0.633 \\

\textbf{Ours}
& \textbf{0.905} & \textbf{0.826}
& \textbf{0.741} & \textbf{0.589}
& \textbf{0.781} & \textbf{0.641}
& \textbf{0.809} & \textbf{0.685} \\

\bottomrule
\end{tabular}

\end{table}

\textbf{Impact of Pixel-Space Modeling.}
The Baseline, a vanilla U-DiT operating directly in pixel space with standard flow matching, establishes a foundational performance level. However, introducing a VAE for latent-space compression (w/ VAE) causes a severe performance drop (e.g., F1 decreases from 0.516 to 0.414 on T-IC13). This result supports our hypothesis that the aggressive downsampling and quantization in VAEs destroy high-frequency micro-artifacts, such as sensor noise and compression grids, that are essential for forensic localization. It therefore justifies our decision to operate directly in pixel space.

\textbf{Efficacy of Forensic-Oriented Components.}
Removing the Multi-modal Forensic Tokenizer (w/o MFT) and relying solely on RGB patches leads to clear degradation, particularly on the challenging OSTF benchmark. This result shows that explicitly encoding noise (SRM) and frequency (DCT) cues is important for guiding restoration-cost estimation. Removing Self-Fingerprint AdaLN (w/o SFA) or Frequency-Gated Attention (w/o FG-MHSA) also degrades performance, although for different reasons. Without SFA, the model becomes less effective at adapting to image-specific forensic backgrounds across different acquisition conditions. Without FG-MHSA, the attention mechanism more easily mixes semantically similar but statistically inconsistent regions, which weakens tampering localization. Our full model (Ours) combines these components and achieves the best overall performance across all benchmarks.

\textbf{Flow Matching vs. Direct Regression.}
To verify that the gains do not simply come from the backbone architecture, we compare our Flow Matching formulation with a direct residual regression baseline under the same F-DiT backbone, the same artifact injection strategy, and the same training pipeline. As shown in Table~\ref{tab:fm_vs_direct_regression}, Flow Matching consistently outperforms direct regression on all three benchmarks by a large margin in the zero-shot setting. The improvement is not uniform across datasets, and is especially pronounced on DocTamper and OSTF, suggesting that the structured time-conditioned supervision provided by Flow Matching is particularly beneficial in more challenging or open-set scenarios. These results indicate that the advantage of our method is not solely architectural, but also comes from the SC-RF training objective itself.

\textbf{Why Flow Matching Despite Single-Step Inference?}
Although our model is trained with Flow Matching, inference only uses the predicted velocity field at $t=0$ as a detector-oriented anomaly signal. We therefore study whether additional ODE steps provide meaningful gains. Table~\ref{tab:inference_steps_ablation} shows that increasing the number of inference steps from 1 to 2 and from 2 to 4 yields only modest improvements, while using 8 steps leads to performance degradation. This result suggests that a single-step estimate already captures the key information needed for tampering localization, namely that authentic regions require little correction whereas tampered regions require large correction magnitude. In practice, 1-step inference provides a strong accuracy-efficiency trade-off, while additional steps substantially increase computational cost but deliver only marginal gains.

\begin{table}[t]
\centering
\caption{Zero-shot ablation on the training objective under the same backbone and the same training pipeline. \textbf{DR} denotes direct residual regression. \textbf{FM} denotes Flow Matching with SC-RF. ``Avg'' denotes the mean over the three benchmarks.}
\label{tab:fm_vs_direct_regression}
\resizebox{\columnwidth}{!}{
\begin{tabular}{lcccccccc}
\toprule
\multirow{2}{*}{Method} 
& \multicolumn{2}{c}{\textbf{T-IC13}}
& \multicolumn{2}{c}{\textbf{DocTamper}}
& \multicolumn{2}{c}{\textbf{OSTF}}
& \multicolumn{2}{c}{\textbf{Avg}}\\
\cmidrule(lr){2-3} \cmidrule(lr){4-5} \cmidrule(lr){6-7} \cmidrule(lr){8-9}
& F1 & IoU & F1 & IoU & F1 & IoU & F1 & IoU \\
\midrule
DR         & 0.822 & 0.698 & 0.614 & 0.443 & 0.654 & 0.486 & 0.697 & 0.542 \\
FM (SC-RF) & \textbf{0.905} & \textbf{0.826} & \textbf{0.741} & \textbf{0.589} & \textbf{0.781} & \textbf{0.641} & \textbf{0.809} & \textbf{0.685} \\
\bottomrule
\end{tabular}
}

\end{table}

\begin{table}[t]
\centering
\caption{Zero-shot ablation on the number of inference steps. \textbf{Step} denotes the number of ODE evaluation steps used at test time. ``Avg'' denotes the mean over the three benchmarks.}
\label{tab:inference_steps_ablation}
\resizebox{\columnwidth}{!}{
\begin{tabular}{lcccccccc}
\toprule
\multirow{2}{*}{Step} 
& \multicolumn{2}{c}{\textbf{T-IC13}}
& \multicolumn{2}{c}{\textbf{DocTamper}}
& \multicolumn{2}{c}{\textbf{OSTF}}
& \multicolumn{2}{c}{\textbf{Avg}}\\
\cmidrule(lr){2-3} \cmidrule(lr){4-5} \cmidrule(lr){6-7} \cmidrule(lr){8-9}
& F1 & IoU & F1 & IoU & F1 & IoU & F1 & IoU \\
\midrule
1-step & 0.905 & 0.826 & 0.741 & 0.589 & 0.781 & 0.641 & 0.809 & 0.685 \\
2-step & 0.910 & 0.835 & 0.748 & 0.597 & 0.788 & 0.650 & 0.815 & 0.694 \\
4-step & \textbf{0.913} & \textbf{0.840} & \textbf{0.752} & \textbf{0.603} & \textbf{0.791} & \textbf{0.654} & \textbf{0.819} & \textbf{0.699} \\
8-step & 0.901 & 0.820 & 0.736 & 0.582 & 0.775 & 0.633 & 0.804 & 0.678 \\
\bottomrule
\end{tabular}
}

\end{table}

% \section{Conclusion}
% In this paper, we propose a generative detector for open-set visual text forensics that departs from conventional discriminative modeling. Instead of learning forgery-specific decision boundaries, our method estimates the local restoration cost required to reduce the discrepancy between a query image and authentic visual-text statistics. Under this detector-oriented formulation, we introduce Sparse-Constraint Rectified Flow to explicitly address the spatial sparsity of tampered regions, and design a texture-centric architecture, Forensic-DiT, together with self-supervised Artifact Injection to preserve and exploit high-frequency forensic traces without paired supervision. Extensive experiments demonstrate that the proposed method achieves strong robustness and generalization, particularly in zero-shot and cross-generator settings. In addition, our auxiliary stress-test analysis suggests that local harmonization can weaken the cues used by existing detectors. Overall, these results indicate that estimating local restoration cost is an effective direction for open-set visual text tampering detection, while its applicability to broader visual manipulation domains remains to be further studied.

\section{Conclusion} We propose a generative detector for open-set visual text forensics that localizes tampering by estimating local restoration costs relative to authentic visual-text statistics. To address sparse anomalies, we introduce Sparse-Constraint Rectified Flow and combine it with the pixel-space Forensic-DiT and self-supervised Artifact Injection to preserve high-frequency forensic traces without paired supervision. Experiments demonstrate strong generalization, particularly in zero-shot settings. Our auxiliary stress test further shows that local harmonization can weaken cues used by existing detectors. These results support restoration-cost estimation as a promising direction for open-set text tampering detection.

\begin{acks}
This work is supported by the National Natural Science Foundation of China
(Grant NO 62376266 and 62406318) and CAAI-Tencent Rhino-Bird
Open Research Fund.
\end{acks}

\bibliographystyle{ACM-Reference-Format}
\balance
\bibliography{references}

\end{document}